\documentclass[10pt,letterpaper]{article}
\usepackage[top=0.85in,left=2.75in,footskip=0.75in]{geometry}

\usepackage{amsmath,amssymb}

\usepackage{changepage}

\usepackage[utf8x]{inputenc}

\usepackage{textcomp,marvosym}

\usepackage{cite}

\usepackage{nameref,hyperref}

\usepackage[right]{lineno}

\usepackage{microtype}
\DisableLigatures[f]{encoding = *, family = * }

\usepackage[table]{xcolor}

\usepackage{array}

\usepackage{hyperref}

\hypersetup{
    bookmarks=true,         
    unicode=false,          
    pdftoolbar=true,        
    pdfmenubar=true,        
    pdffitwindow=false,     
    pdfstartview={FitH},    
    pdftitle={My title},    
    pdfauthor={Author},     
    pdfsubject={Subject},   
    pdfcreator={Creator},   
    pdfproducer={Producer}, 
    pdfkeywords={keywords}, 
    pdfnewwindow=true,      
    colorlinks=true,       
    linkcolor=blue,          
    citecolor=blue,        
    filecolor=magenta,      
    urlcolor=cyan           
}

\newcolumntype{+}{!{\vrule width 2pt}}

\newlength\savedwidth

\raggedright
\usepackage[aboveskip=1pt,labelfont=bf,labelsep=period,justification=raggedright,singlelinecheck=off]{caption}

\makeatletter
\renewcommand{\@biblabel}[1]{\quad#1.}
\makeatother

\usepackage{lastpage,fancyhdr,graphicx}
\usepackage{epstopdf}
\fancyheadoffset[L]{2.25in}
\fancyfootoffset[L]{2.25in}
\begin{document}
\vspace*{0.2in}

\begin{flushleft}
{\Large
\textbf\newline{   {Mediating artificial intelligence developments  through  negative and positive  incentives}} 
}
\newline
\\
The Anh Han\textsuperscript{1*},
 Lu\'is Moniz Pereira\textsuperscript{2},
Tom Lenaerts\textsuperscript{3,4},
Francisco C. Santos\textsuperscript{5}
\\
\bigskip
\textbf{\textsuperscript{1}}School of Computing and Digital Technologies,  Teesside University, Middlesbrough, UK TS1 3BA
\\
\textbf{\textsuperscript{2}}NOVA Laboratory for Computer Science and Informatics (NOVA LINCS), 
  Universidade Nova de Lisboa, 
        2829-516 Caparica, Portugal
\\
\textbf{\textsuperscript{3}}Machine Learning Group, Universit{\'e} Libre de Bruxelles, Boulevard du Triomphe CP212, Brussels, Belgium
\\
\textbf{\textsuperscript{4}}Artificial Intelligence Lab, Vrije Universiteit Brussel, 1050 Brussels, Belgium
\\
\textbf{\textsuperscript{5}}INESC-ID and Instituto Superior Tecnico, Universidade de Lisboa,  IST-Taguspark, 2744-016, Porto, Salvo, Portugal
\\
\bigskip

%
%





* Corresponding author: The Anh Han (t.han@tees.ac.uk)

\end{flushleft}
\section*{Abstract}
The field of Artificial Intelligence (AI)  is going through a period of great expectations, introducing a certain level of anxiety in research, business and also policy. This anxiety is further energised by an AI race narrative that makes people believe they might be missing out. Whether real or not,  a belief in this narrative may be detrimental as some stake-holders will feel obliged to cut corners on safety precautions, or ignore  societal consequences just to ``win".  Starting from a baseline model that describes a broad class of  technology races where winners draw a significant benefit compared to others (such as AI advances, patent race, pharmaceutical technologies), we investigate here how positive (rewards) and negative (punishments) incentives may beneficially  influence  the outcomes. We uncover conditions in which punishment is  either capable of reducing the development speed of unsafe participants or has the capacity to reduce innovation through over-regulation. Alternatively, we show that, in several scenarios, rewarding those that follow safety measures may increase the development speed while ensuring safe choices. Moreover, in  {the latter} regimes, rewards do not suffer from the issue of over-regulation as is the case for punishment. Overall, our findings provide valuable insights into the nature and kinds of regulatory actions most suitable to improve safety compliance in the contexts of both smooth and sudden technological shifts.



\section*{Introduction}
With the current business and governmental anxiety about AI and the promises made about the impact of AI technology, there is a risk for stake-holders to cut corners, preferring rapid deployment of their AI technology over an adherence to safety and ethical procedures, or {a willingness to examine} their societal impact \cite{armstrong2016racing,cave2018ai,goldai2}.

Agreements and regulations for safety and ethics can be enacted by involved parties so as to ensure their compliance concerning mutually adopted standards and norms \cite{shulman2009arms}. However,  experience with a spate of international treaties, like those of climate change, timber, and fisheries  agreements \cite{barrett2016coordination,cherry2013enforcing,nesse2001evolution} has shown, the autonomy and sovereignty of the parties involved will make monitoring and compliance enforcement difficult (if not impossible). Therefore, for all to enjoy the benefits provided by  safe, ethical and trustworthy AI, it is crucial to    {design and} impose appropriate incentivising strategies in order to ensure mutual benefits and safety-compliance from all sides involved.
Given these concerns, many calls for developing efficient forms of regulation  have been made \cite{baum2017promotion,cave2018ai,taddeo2018regulate}. 

In this paper, we aim to understand how different forms of incentives can be efficiently used to influence safety decision making within a development race for  domain supremacy through AI (DSAI), resorting to population dynamics and Evolutionary Game Theory (EGT) \cite{maynard-smith:1982to,nowak:2006bo,key:Sigmund_selfishnes}. 
Although  AI development is used here to frame the model and to discuss the results, both model and conclusions may easily be adopted for other technology races, especially where a winner-takes-all situation occurs  \cite{denicolo2010winner,campart2014technological,lemley2012myth}.

We posit that it requires time to reach DSAI, modelling this by a number of development steps or technological advancement rounds \cite{han2019modelling}. In each round the development teams (or players) need to choose between one of two strategic options:  to follow safety precautions (the SAFE action) or ignore safety precautions (the  UNSAFE action). Because it takes more time and more effort to comply with precautionary requirements, playing SAFE is not just costlier, but  implies slower development speed too, compared to  playing UNSAFE. We consequently assume that to play SAFE involves paying a cost $c > 0$, while playing UNSAFE  costs nothing  ($c = 0$).  Moreover, the development  speed of playing UNSAFE is $s > 1$  whilst the speed of playing SAFE is normalised to $s = 1$. 
The interaction is iterated until one or more teams establish DSAI, which occurs probabilistically, i.e.  the model assumes, upon completion of each round, that there is a probability $\omega$ that another development round is required to reach DSAI---which results in an average number $W = (1 - \omega)^{-1}$ of rounds per competition/race \cite{key:Sigmund_selfishnes}. We thus do not make any assumption about the time required to reach DSAI in a given domain. Yet once the race ends, a large benefit or prize $B$ is acquired that is  shared amongst those  reaching the target simultaneously.  

The DSAI model further assumes that a development setback or disaster might  occur, with a probability assumed to increase with the number of    occasions  the safety requirements have been omitted by the winning team(s) at each round. Although many potential AI disaster scenarios have been sketched \cite{armstrong2016racing,pamlin2015global}, the uncertainties in accurately predicting these outcomes  have been shown to be high.
 When such a disaster occurs, the risk-taking participant loses all its accumulated benefits, which is denoted by $p_r$, the risk probability of such a disaster occurring when no safety precaution is followed  (see Materials  and Methods section for further details). 

\begin{figure}
\centering
\includegraphics[width=0.6\linewidth]{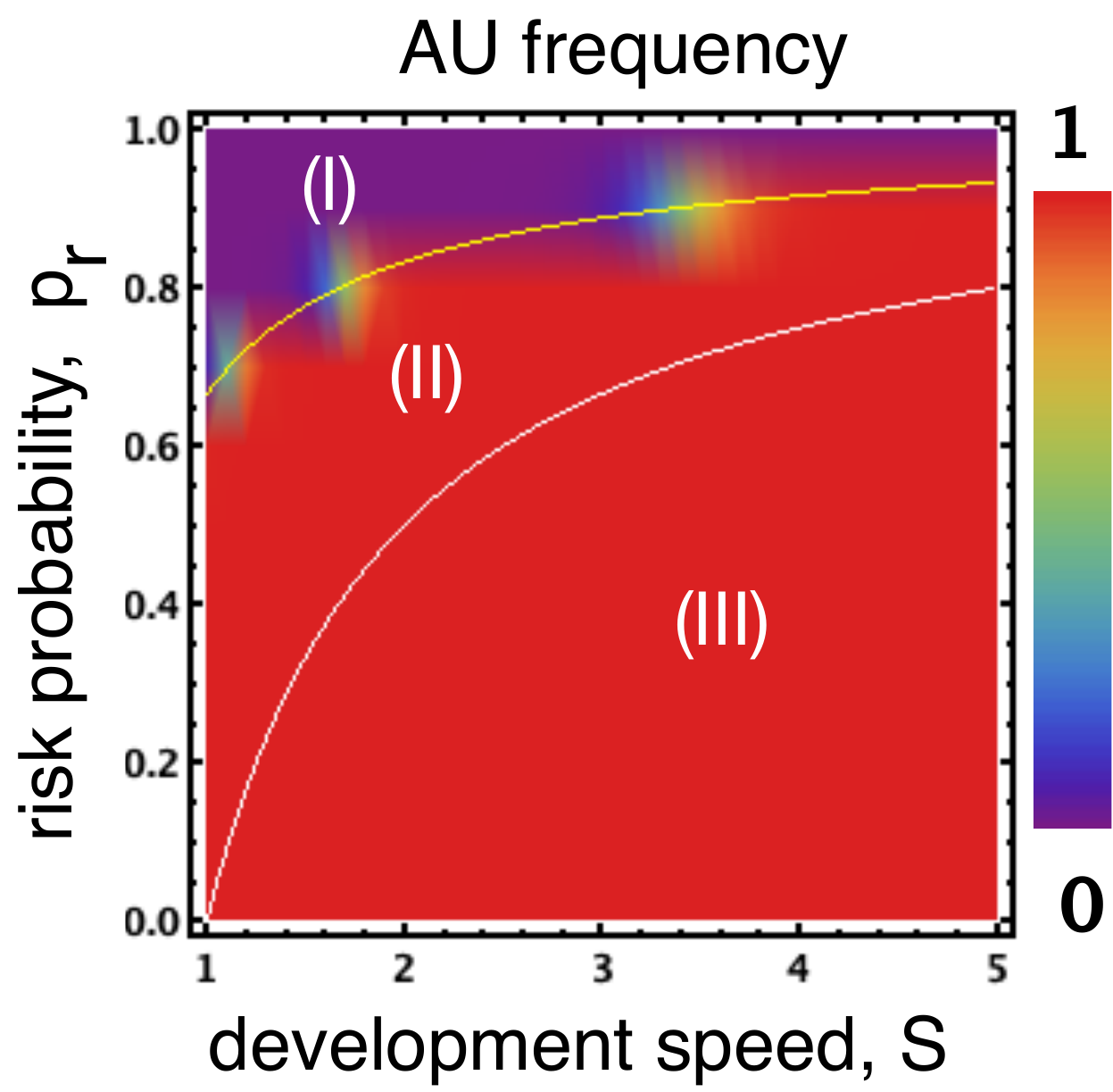} 
\caption{\textbf{Frequency of AU in a population of AU and AS.} Region (\textbf{II}): The two solid lines inside the plots delineate the boundaries $p_r \in [1-1/s, 1-1/(3s)]$  where safety    {compliance} is the preferred collective outcome yet evolution selects unsafe development. Regions (\textbf{I}) and (\textbf{III}) display where safe (respectively, unsafe)    {development is not only the preferred collective outcome but also the one selected by evolution}. Parameters:  $b = 4$, $c = 1$, $W = 100$, $B = 10^4$, $\beta = 0.01$,  $Z = 100$. 
}
\label{fig:panel_no_punishment}
\end{figure}

As shown in \cite{han2019modelling}, when the time-scale of reaching the target is short, such that the average benefit over all the development rounds, i.e. $B/W$, is significantly larger compared to the intermediate benefit $b$ obtained in every round, there is a large parameter space where societal interest is in conflict with the personal one: unsafe behaviour is dominant despite the fact that safe development would lead to a greater social welfare (see region \textbf{II} in Figure \ref{fig:panel_no_punishment} and Supporting Information (SI)  for details). The reason is that, those who completely ignore safety precautions can always achieve the big prize $B$ when playing against safe participants. 
 The two other zones, i.e. region \textbf{I} and region \textbf{III}  in Figure \ref{fig:panel_no_punishment}, do not suffer from a dilemma between individual and group benefits as is the case for region  \textbf{II}.  Whereas in region \textbf{I} safe development is preferred due to excessively high risks, region \textbf{III} prefers unsafe, risk taking behaviour, both from an individual and societal perspective.

From a regulatory perspective, only region \textbf{II} requires additional measures that ensure or enhance safe and globally beneficial  outcomes, avoiding any potential disaster. Large-scale surveys and expert analysis of the beliefs and predictions about the progress in AI, indicate that the perceived time-scale for  supremacy across domains through AI as well as regions is highly diverse  \cite{armstrong2014errors,grace2018will}. 
Also note that despite focusing on DSAI in this paper, the proposed model is generally applicable to any kind of long-term competing situations such as technological innovation development and patent racing where there is a significant advantage (i.e. large $B$) to be achieved by reaching an important target first \cite{denicolo2010winner,campart2014technological,lemley2012myth}. Other  domains include pharmaceutical development where firms could try to cut corners by not following safe clinical trial protocols in an effort to be the first to develop a pharmaceutical produce (i.e.  a cure for cancer), in order to take the highest possible share of the market benefit  \cite{abbott2009global}; Besides tremendous  economic advantage, a winner of a vaccine race such as for Covid-19 treatment, can also gain significant political and reputation influence \cite{burrell2020covid}. 

In this paper, we explore whether and how incentives such as reward and punishment can help in avoiding disasters and generate a wide benefit of AI-based solutions. Namely, players can attempt to prevent others  from moving as fast as they want (i.e., an elementary form of punishment of wrong-doers) or help others to speed up  their development (rewarding  right-doers), at a given cost. Slowing down unsafe participants can be obtained by reporting misconduct to authorities and media, or by refusal to share and collaborate with companies not following the same deontological principles. Similarly, rewards can correspond to support, exchange of knowledge, staff, etc. of safety conscious participants. Note that reasons for intervening with the development speed of competitors may also be nefarious, e.g. cyber-attacks, in order to get a speed advantage. The current work only considers interventions by safe players as a result of the unsafe behaviour of co-players. We show that both negative and positive incentives can be efficient and naturally self-organize (even when costly). However, we also show that such incentives should be carefully introduced,  as they can have negative effects otherwise. To this end, we identify the conditions under which positive and negative incentives are conducive to desired collective outcomes. 

\subsection*{Related Work}
Although there have been a number of proposals and debates on how to avert, regulate, or mediate a race for technological supremacy  \cite{baum2017promotion,cave2018ai,geist2016s,shulman2009arms,vinuesa2019role,taddeo2018regulate,askell2019role}, few  formal modelling studies were proposed \cite{armstrong2016racing,han2019modelling}. The current paper takes the next step, further filling this gap. Namely, it will resort to Evolutionary Game Theory (EGT) methods to investigate   how  positive and negative incentives can improve the outcomes of DSAI and, more generally, a broad class of innovation race dynamics.  

Incentives such as  punishment and rewards have been shown to provide important mechanisms to promote the emergence of  positive behaviour (such as cooperation and fairness) in the context of social dilemmas  \cite{key:fehr2002,Sigmund2001PNAS,boyd2010coordinated,sigmundinstitutions,Han2016AAAI,hilbe2012emergence,szolnoki2013correlation,gois2019reward,Hanijcai2018,chen2015first,garcia2019evolution}.
Notwithstanding, all existing modelling approaches to AI governance \cite{armstrong2016racing,han2019modelling} do not
 study how incentives can be used to enhance  safety compliance. Moreover, there have been {incentive-modelling} studies addressing  other kinds of risk, such as climate change and nuclear war{, see e.g.} \cite{vasconcelos2013bottom,gois2019reward,baliga2004arms}.    {Following from an analysis of several  large global catastrophic risks  \cite{pamlin2015global}, it has been shown that the race for domain supremacy through AI and its related risks are rather unique}. Analyses of climate change disasters primarily focus on participants'  unwillingness   to take upon themselves some personal cost for a desired collective target, and  implies a collective risk for all  parties involved \cite{gois2019reward}. In contrast, in a race to become leader in a particular AI application domain, the winner(s) will    {extract} significant  advantage    {relative to that of} others.    {More importantly, this AI risk}  is also more directed    {towards individual developers or users} than collective ones. 

\section*{Materials and methods}
\label{section:models and methods}
\subsection*{DSAIR model definition}
Let us depart from the innovation race or domain supremacy through AI race (DSAIR) model  developed in \cite{han2019modelling}. We adopt a  two-player repeated game, consisting of, on average, $W$ rounds. 
At each    {development}  round,  players can collect benefits from their intermediate AI products, subject to whether they choose  playing SAFE or UNSAFE. By assuming some fixed benefit, $b$, resulting from the AI market, the teams share this benefit in proportion to their development speed. 
Hence, for every round of the race, we can write, with respect to the row player $i$, a payoff matrix denoted by $\Pi$, where each entry is represented by $\Pi_{ij}$ (with $j$ corresponding to a column),  as follows  
{\begin{equation}
\Pi =  \bordermatrix{~ & \textit{SAFE} & \textit{UNSAFE}\cr
                  \textit{SAFE} & -c + \frac{b}{2} &-c +  \frac{b}{s+1}      \cr
                  \textit{UNSAFE} &  \frac{s b}{s+1}   & \frac{b}{2}   \cr
                 }.
\end{equation}
}
The payoff matrix can be explained as follows. First of all, whenever two SAFE players interact, each will pay the cost $c$ and  share the resulting benefit $b$. {Differently, when two UNSAFE players interact, each will share the  benefit $b$ without having to pay $c$}. 
{When a SAFE player interacts with an UNSAFE player, the SAFE one pays a cost $c$} {and  receives a (smaller) part $b/(s+1)$ of the benefit $b$, while the UNSAFE one obtains the larger part $sb/(s+1)$ without having to pay $c$}.  Note that $\Pi$ is a simplification of the matrix defined in \cite{han2019modelling} since it was shown that the parameters defined here are sufficient to explain the results in the current time-scale.

 We will analyse evolutionary outcomes of safety behaviour within a  well-mixed, finite  population consisting of  $Z$ players, who repeatedly interact with each other  in the AI development process. They will  adopt one of the  following  two strategies \cite{han2019modelling}: 
\begin{itemize} 
	\item \textbf{AS}: always complies with safety precaution, playing SAFE in all the rounds.  
	\item \textbf{AU}: never  complies with safety precaution, playing UNSAFE in all the rounds. 
\end{itemize} 
The payoff matrix defining averaged payoffs for {AU vs AS}    {is given by}    
\begin{equation}
 \bordermatrix{~ & \textit{AS} & \textit{AU} \cr
                  \textit{AS} & \frac{B}{2W} +\Pi_{11} & \Pi_{12}    \cr
                  \textit{AU} &  p \left(\frac{sB}{W} + \Pi_{21}\right)   &  p \left(\frac{sB}{2W} +\Pi_{22}\right)  \cr 
                 },
\end{equation}
where, solely with the purpose of presentation, we denote $p = 1 - p_r$.

As was shown in \cite{han2019modelling} by considering when AU is risk-dominant against AS,  three different regions can be identified in the parameter space $s$-$p_r$ (see Figure \ref{fig:panel_no_punishment}), details are provided in SI): (\textbf{I})  when $ p_r > 1-\frac{1}{3s}$, AU is risk-dominated by  AS:   safety {compliance} is both the preferred collective outcome and selected by evolution; (\textbf{II}) when $1-\frac{1}{3s} > p_r > 1-\frac{1}{s}$:  even though it is more desirable to ensure  safety    {compliance}  as the collective outcome,    {social learning dynamics} would    {lead}  the population to the state    {wherein the safety precaution} is mostly ignored;  (\textbf{III}) when $p_r <  1-\frac{1}{s}$ (AU is risk-dominant against  AS),  then unsafe development is both    {preferred collectively and} selected by    {social learning dynamics}. 

It is worthy of note that  adding a  conditional strategy (that, for instance, plays SAFE in the first round and thereafter adopts    {the same move its co-player used on} the previous round) does not influence the dynamics or improve  safe outcomes (see details in SI). This is contrary to the prevalent models of direct reciprocity  in  the repeated social dilemmas context \cite{key:Sigmund_selfishnes,van2012emergence,key:HanetalAlife}.  
Therefore, additional measures need to be put in place for driving the race dynamics towards a more beneficial outcome. To this end, we came to explore in this work the effects of negative (sanctions) and positive (rewards) incentives. 

\subsection*{Punishment and reward in innovation races}

\begin{figure}[h!]
\centering
\includegraphics[width=0.8\linewidth]{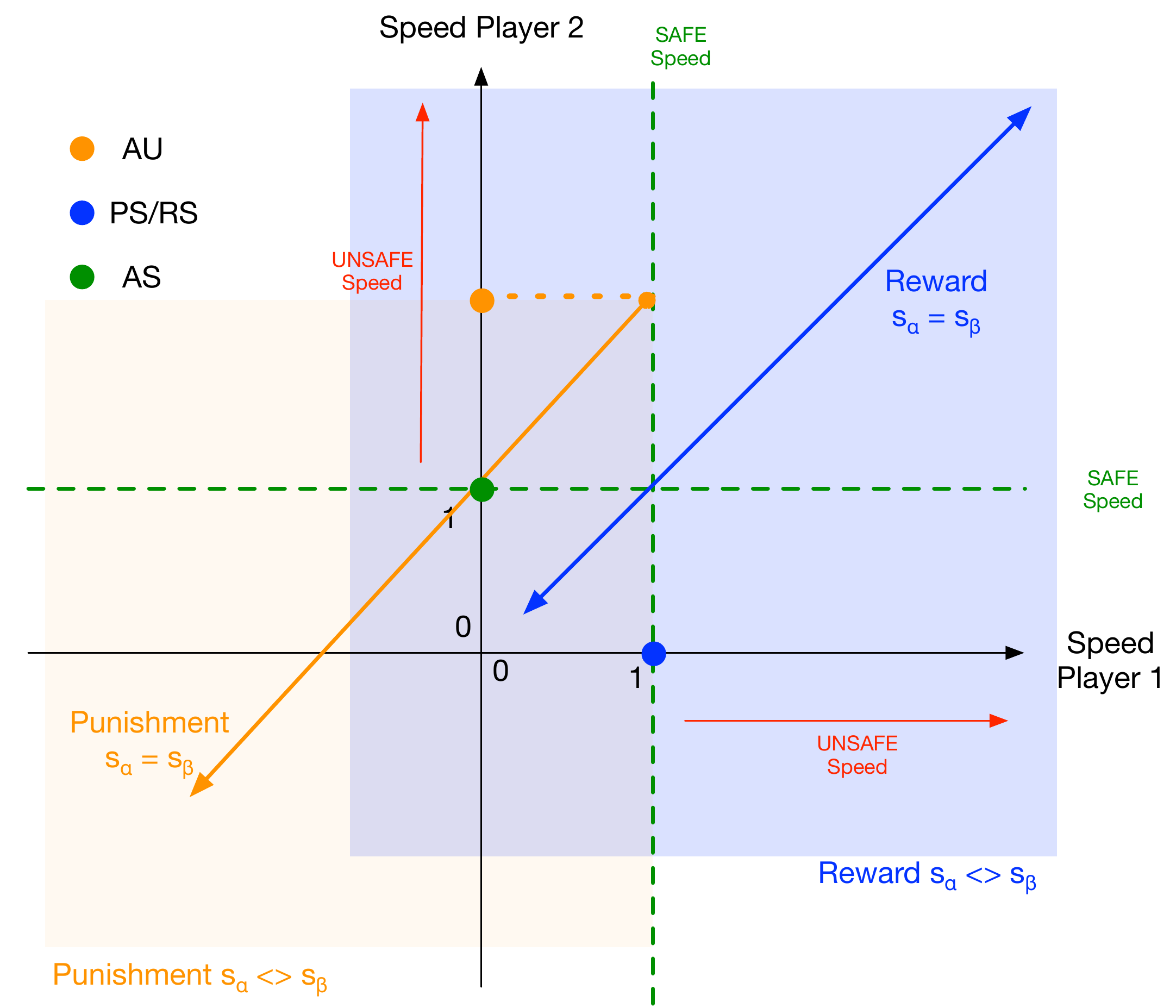} 
\caption{\textbf{Effect of positive and negative incentives on players' speed.} On the on hand, when player 1 is of type PS (blue circle on x-axis), i.e. sanctioning unsafe actions, it reduces the future speed of player 2 when she is of type AU (orange circle on the y-axis), while paying a speed cost, possibly equivalent to the reduction in speed that the AU player is experiencing (orange line).  In general the reduction of speeds of player 1 and 2 fall into the area marked by the orange rectangle.  On the other hand, when player 1 is of type RS (blue circle on x-axis), i.e. rewarding safe actions, it increases the speed of player 2 (green circle on y-axis), while paying a speed cost that reduces the RS player's speed. Differently from before, the speed effect is in opposing directions for the two players. The blue rectangle marks the area of the speed of player 1 and player 2. In the analysis in the paper, first the case of equal speed effects is considered (lines) before analysing different speed effects (rectangles) between both players. 
}
\label{fig:incentives_effect}
\end{figure}

Given the DSAIR model one can now introduce incentives that affect the development speed of the players.  These incentives reduce or increase the speed of development of a player as this is the key factor in gaining $b$ as well as $B$ once the game ends \cite{han2019modelling}. While there are many ways to incorporate them, we assume here  a minimal model where the effect on speed is constant and fixed over time, hence not cumulative with the number of unsafe or safe actions of the co-player. Given this constant assumption, a negative incentive reduces the speed of a co-player taking an UNSAFE action to a lower but constant speed-level. Similarly, a positive incentive increases the speed of a co-player that took a safe action to a fixed higher speed-level. In both cases these incentives are attributed in the next round, after observing the UNSAFE or SAFE action respectively. Moreover, both positive and negative incentives are considered to be costly, meaning that the strategy that awards them will reduce its own speed by providing the incentive.    Given these assumptions the following two strategies are studied in relation to the AS and AU strategies defined earlier: 

\begin{itemize}
\item A strategy PS that always plays SAFE but will sanction the co-player after she has played UNSAFE in the previous round. The punishment by PS imposes a  reduction $s_\beta$ on the opponent's speed as well as a reduction $s_\alpha$ on her own speed (see Figure \ref{fig:incentives_effect}, orange line/area).

\item A strategy RS that always chooses the SAFE action and will reward a SAFE action of a co-player by increasing her speed with $s_\beta$  while paying a cost $s_\alpha$ on her own speed  (see Figure \ref{fig:incentives_effect}, blue line/area).  
\end{itemize}

The analysis performed in the Results section aims to show whether having PS or/and RS in the population leads to more societal welfare in the region (\textbf{II}), where there is a conflict between individual and societal interests.  The methods used in this analysis are discussed in the next section.

\subsection*{Evolutionary Dynamics    {for} Finite Populations}

\begin{figure*}
\centering
\includegraphics[width=\linewidth]{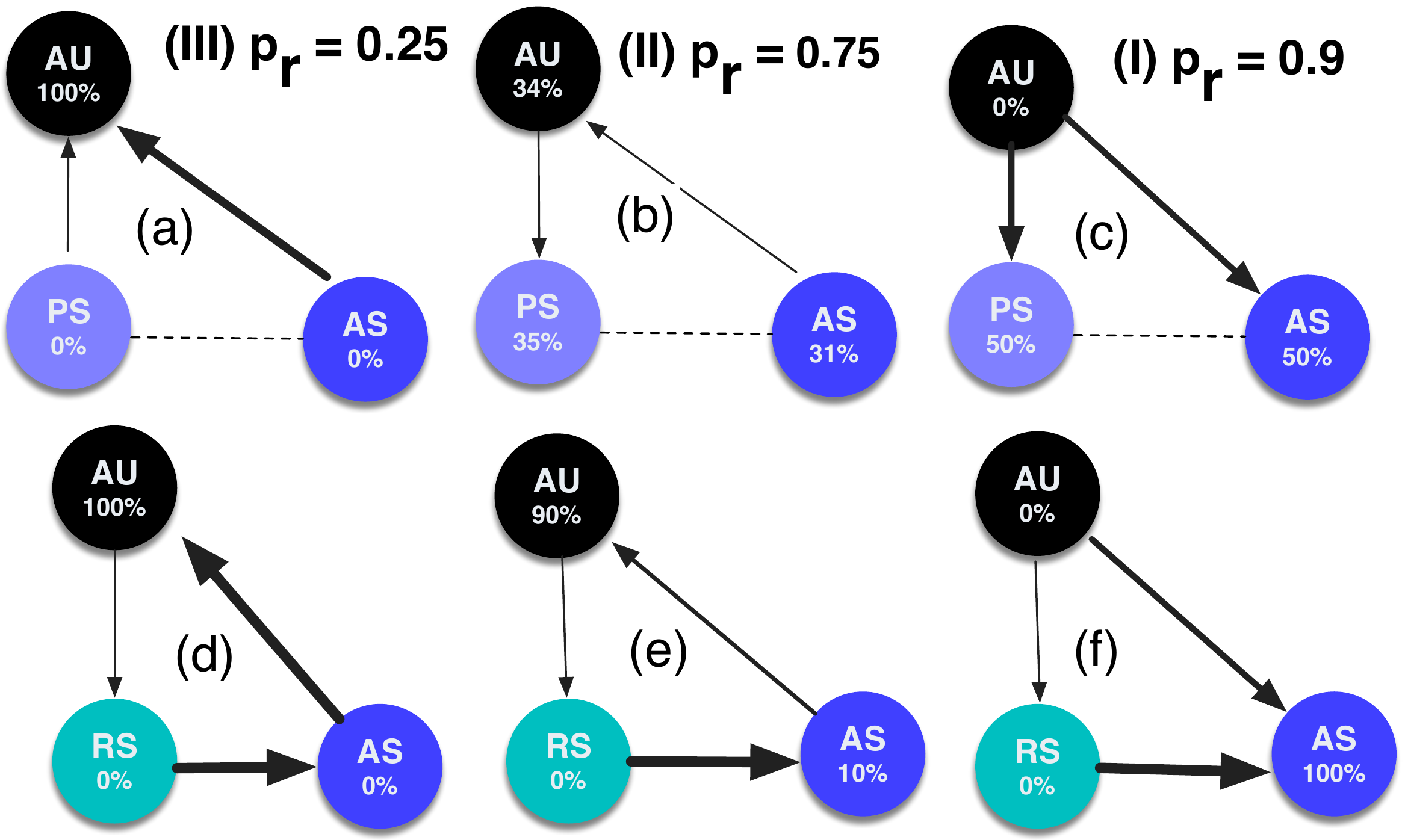}
\caption{Transitions and stationary distributions  in a population of three strategies AU, AS, with either  PS (top row) or RS (bottom row), for three regions. Only stronger transitions are shown for clarity. Dashed lines denote  neutral transitions. Parameters: $s_\alpha = s_\beta  = 1.0$, $c = 1$, $b = 4$, $W = 100$,  $B = 10000$, $\beta = 0.01$, $Z = 100$. 
}
\label{fig:panel_markov}
\end{figure*}

We employ EGT methods for finite populations \cite{key:Sigmund_selfishnes,traulsen2006,hindersin2019computation}, whether in the analytical or numerical results obtained here.   Within such a setting, the  players'  payoffs stand for their \emph{fitness} or social \emph{success}, and  social learning shapes the evolutionary dynamics, according to which the  most successful players will more often tend to be imitated  by other players. Social learning is herein modeled utilising  the so-called pairwise comparison rule \cite{traulsen2006},  assuming  that a player $A$ with fitness $f_A$ adopts the strategy of another player $B$ with fitness $f_B$ with probability    {assigned} by the Fermi function, 
$P_{A,B}=\left(1 + e^{-\beta(f_B-f_A)}\right)^{-1}$,  
where  $\beta$ conveniently describes the intensity of selection. 
The long-term frequency of each and every strategy in a population where several of them are in co-presence, can be computed simply by calculating  the stationary distribution of a Markov chain whose states represent those strategies. 
In the absence of    {behavioural} exploration  or mutations, end states of evolution inevitably are monomorphic.    {That is, whenever}
 such a state is reached, it cannot be escaped via imitation. Thus, we further assume that, with some mutation probability,  an agent can     {freely explore its behavioural space (in our case, consisting of two actions,  SAFE and UNSAFE), randomly adopts an action  therein}.  At the limit of    {a} small mutation    {probability},    {the population consists of at most two strategies at any time. Consequently,  the social dynamics can be described using a Markov Chain, where its state } represents a monomorphic population and its transition probabilities are given by the fixation probability of a single mutant \cite{key:imhof2005,key:novaknature2004}.    {The  Markov Chain's stationary distribution  describes the time average}  the population spends in each of the monomorphic end states (see already the examples in Figure \ref{fig:panel_markov} for illustration). 

 Denote by $\pi_{X,Y}$ the payoff a strategist X obtains in a pairwise interaction with strategist $Y$ (defined in the payoff matrices). Suppose there exist at most two strategies in the population, say, $k$ agents using strategy A ($0 \leq k \leq Z$)  and $(Z-k)$ agents using strategies B. Thus, the (average) payoff of the agent that uses  A and B can be written as follows, respectively, 
\begin{equation} 
\label{eq:PayoffA}
\begin{split} 
\Pi_A(k) &=\frac{(k-1)\pi_{A,A} + (Z-k)\pi_{A,B}}{Z-1},\\
\Pi_B(k) &=\frac{k\pi_{B,A} + (Z-k-1)\pi_{B,B}}{Z-1}.
\end{split}
\end{equation} 
Now, in each time step, the probability    {of change by $\pm$1 of a number of  $k$ agents using strategy A} can be specified as\cite{traulsen2006} 
\begin{equation} 
T^{\pm}(k) = \frac{Z-k}{Z} \frac{k}{Z} \left[1 + e^{\mp\beta[\Pi_A(k) - \Pi_B(k)]}\right]^{-1}.
\end{equation}
The fixation probability of a single mutant    {adopting} A, in a population of $(Z-1)$ agents    {adopting} B, is specified by \cite{traulsen2006,key:novaknature2004}
\begin{equation} 
\label{eq:fixprob} 
\rho_{B,A} = \left(1 + \sum_{i = 1}^{Z-1} \prod_{j = 1}^i \frac{T^-(j)}{T^+(j)}\right)^{-1}.
\end{equation} 
When considering a set  $\{1,...,s\}$ of distinct strategies, these fixation probabilities determine the Markov Chain transition matrix $M = \{T_{ij}\}_{i,j = 1}^s$, with $T_{ij, j \neq i} = \rho_{ji}/(s-1)$ and  $T_{ii} = 1 - \sum^{s}_{j = 1, j \neq i} T_{ij}$. The normalized eigenvector    {of the transposed of $M$ associated with the eigenvalue 1} provides the above described  stationary distribution  \cite{key:imhof2005}, which defines the relative time the population spends while adopting each of the strategies. 
\paragraph{Risk-dominance} An important    {approach} for comparing  two strategies A and B is that of in which direction the transition is stronger or more probable, that of an A mutant  fixating in a population of agents employing B, $\rho_{B,A}$,  or that of a B mutant fixating in the population of agents employing A, $\rho_{A,B}$.    {In the limit, for large population size (large $Z$), this condition can be simplified to} \cite{key:Sigmund_selfishnes}  
\begin{equation} 
\label{eq:risk_dominance Equation}
\pi_{A,A} + \pi_{A,B} > \pi_{B,A} +  \pi_{B,B}.
\end{equation}

\begin{figure*}
\centering
\includegraphics[width=\linewidth]{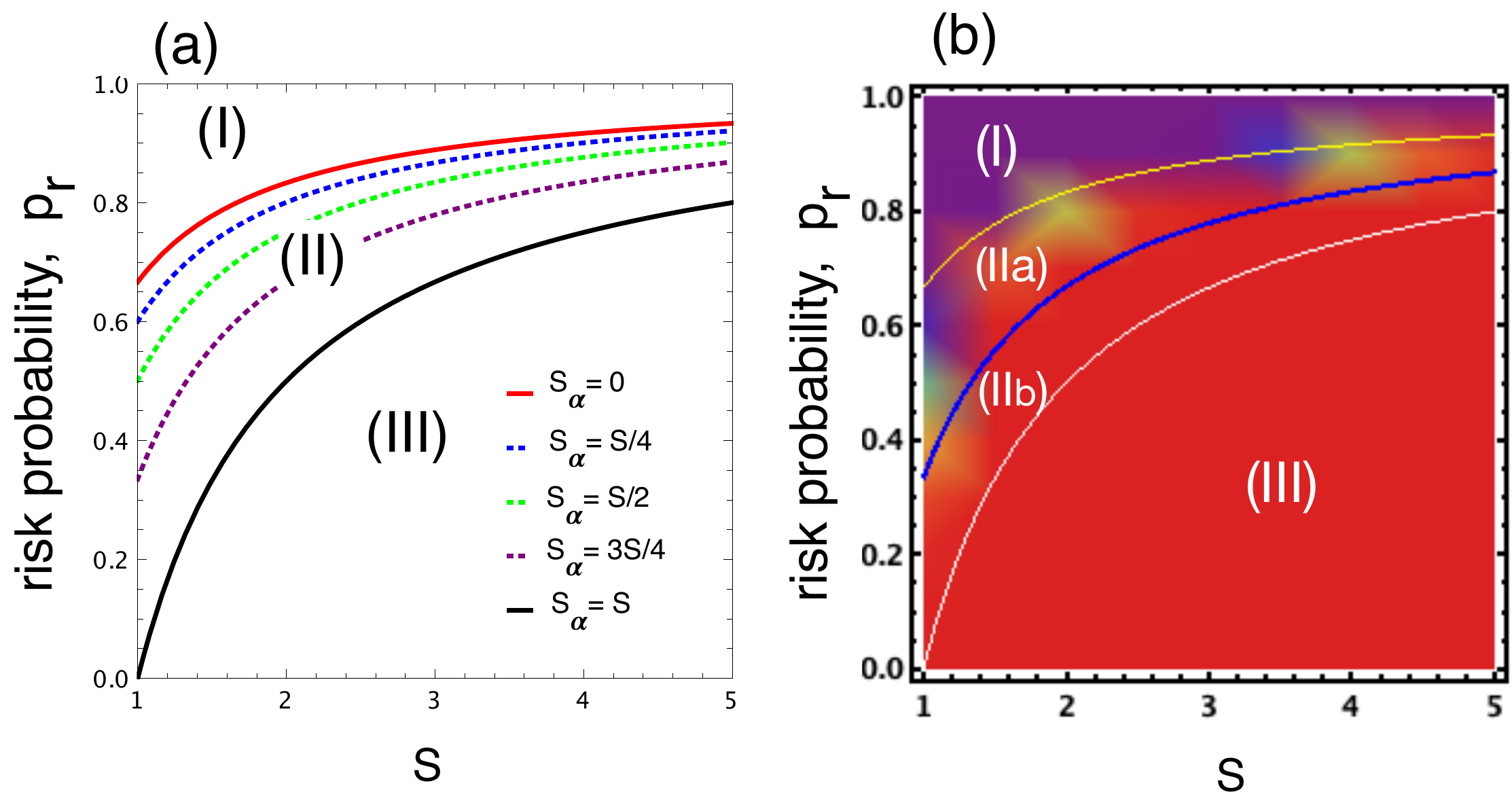}
\caption{(a)  Risk-dominant condition of PS against AU, as defined  in Equation \ref{eq:risk_dom_social_pun}, for different ratio $s_\alpha/s$.  The two solid lines correspond to when the ratio is 0 and 1, corresponding to the boundaries $p_r \in [1-1/s, 1-1/(3s)]$. The larger the ratio the smaller the  Region (\textbf{II}) (between this line and the black line) is decreased, which disappears when $s_\alpha = s$.  Panel (b): frequency of AU in a population of AS, AU, and PS (for $s_\alpha = 3s/4$). Region (\textbf{II}) is split into two (\textbf{IIa}) and (\textbf{IIb}) where PS is now also be preferred to AU in the first one.   Parameters: $b = 4$, $c = 1$,  $W = 100$,   $B = 10000$, $\beta = 0.01$, $Z = 100$. 
}
\label{fig:risk_dom_social_punishment_PSvsAU}
\end{figure*}

\begin{figure*}
\centering
\includegraphics[width=\linewidth]{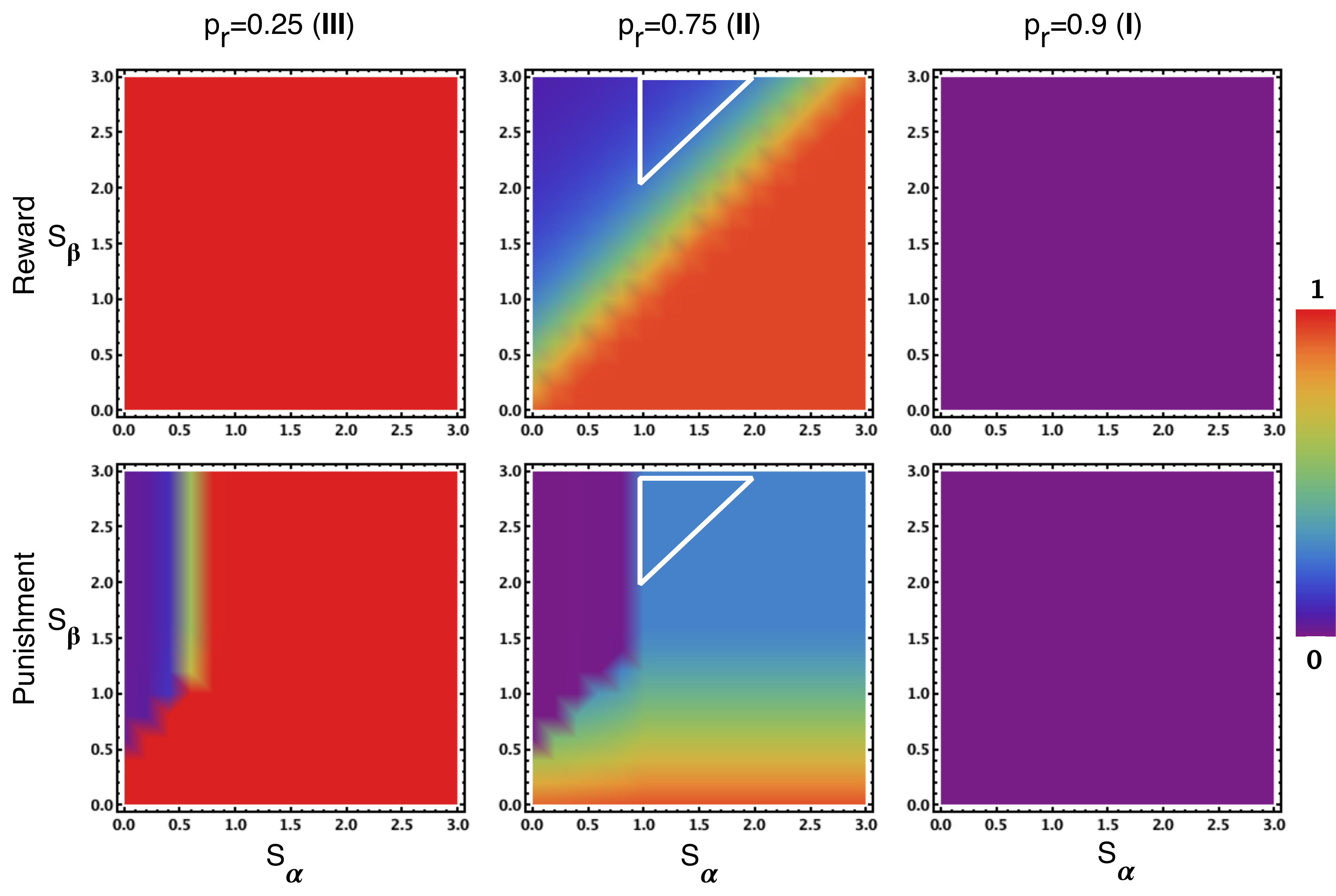}
\caption{\textbf{AU Frequency: Reward (top row) vs punishment (bottom row) for varying $s_\alpha$ and $s_\beta$}, for three regions. In (\textbf{I}), both lead to no AU, as desired. In (\textbf{II}), punishment is more efficient except for when reward is rather costly but highly cost-efficient (the areas inside the white triangles). It is noteworthy that RS has very low frequency in all cases, as it catalyses the success of AS. In (\textbf{III}), {RS always leads to the desired outcome of high AU frequency, while PS might lead to an undesired result of a reduced AU frequency (over-regulation) when highly efficient (non-red area).}  Parameters: $b = 4$, $c = 1$, $W = 100$,  
 $B = 10000$,  $s = 1.5$, $\beta = 0.01$, population size, $Z = 100$. }
\label{fig:reward_vs_punishment_countour_vary_sAlpha_sBeta}
\end{figure*}

\section*{Results}
\subsection*{Negative incentives are a double-edged sword}

As explained in Methods  PS reduces the speed of an AU player from $s$ to $s-s_\beta$, while reducing its own speed from $1$  (since it plays always SAFE) to $1-s_\alpha$.  Hence one can define $s^\prime =1- s_\alpha$ as the new speed for PS and $s^{\prime\prime} =  s- s_\beta $ as the new speed for AU. Depending on the values of $s_\alpha$ and $s_\beta$, these speeds may also be zero or even negative, which represent situations where no progress is being made or where punishment even destroys existing development, respectively. In the following we consider these situations in two different ways. First, a theoretical analysis is performed for the situation where $s_\beta = s_\alpha$.  Second, this assumption is relaxed and a numerical study of the generalised case is provided.  

There are two scenarios to consider when $s_\beta = s_\alpha$: (i) when $s_{\alpha} \geq s$ and (ii) when it is not.  In scenario (i),  $s^{\prime}$ and $s^{\prime\prime}$ are non-positive, resulting in an infinite number of rounds since the target can never be reached.  The average payoffs of PS and AU when playing against each other are thus  $-c$ and $0$, respectively  (assuming that when a team's development speed is {non-positive}, its intermediate benefit, $b$,  is zero).  
The condition for PS to be risk-dominant against AU (see Equation \ref{eq:risk_dominance Equation} in Methods, and noting that the payoff of PS against another PS is the same as that of AS against another AS) reads 
$$ (1-p_r) \left(\frac{sB}{2W} +\Pi_{22}\right)   < \frac{B}{2W} + \Pi_{11} - c.$$
For sufficiently large $B$ (fixing $W$), this  condition is reduced to, $p_r > 1 - 1/s$. That is,   PS is risk-dominant against AU for the whole region (\textbf{II}), thereby ensuring that safe behaviour becomes promoted in that dilemma region. 

Considering  the second case in scenario (ii), where  $s_\alpha < s$, the game is repeated for $\frac{W-s}{s-s_\alpha} + 1 = \frac{W-s_\alpha}{s-s_\alpha}$ rounds, which we {denote here by $r$. Hence, the payoffs of PS and AU when playing with each other are given by, respectively 
$$  \frac{1}{r} \left( \pi_{12} + (r-1)   \pi^\prime_{12} \right),  $$
$$  \frac{p}{r} \left(B +  \pi_{21} + (r-1)   \pi^\prime_{21} \right),  $$
  where $$\pi^\prime_{12}  = \begin{cases} -c &\mbox{if } s > s_\alpha \geq 1  \\
 -c + \frac{(1-s_\alpha)b}{s+1 - 2 s_\alpha}   &\mbox{if }  s_\alpha < 1
 \end{cases},  $$ and 
 $$\pi^\prime_{21}  = \begin{cases}  b &\mbox{if } s > s_\alpha \geq 1  \\
 \frac{(s-s_\alpha)b}{s+1 - 2 s_\alpha}     &\mbox{if }  s_\alpha < 1
 \end{cases}.  $$
Thus, for sufficiently large $B$,   PS is risk dominant against AU when 
$$ p \frac{sB}{2W} +   \frac{p}{r} B < \frac{B}{2W}, $$
which is simplified to:
\begin{equation} 
\label{eq:risk_dom_social_pun}
p_r > 1 - \frac{1}{s + 2W r}.
\end{equation} 
This condition is easier to achieve for small{er} $r$.  
Since $r$ is an {increasing}  function of $s_\alpha$, to optimise the safety outcome, the highest possible $s_\alpha$ should be adopted, i.e. the strongest possible effort in slowing down the opponent should be made.  Figure \ref{fig:risk_dom_social_punishment_PSvsAU}a shows the condition for different values of $s_\alpha$ in relation to $s$ (fixing the ratio $s_\alpha/s$).  Numerical results in Figure \ref{fig:risk_dom_social_punishment_PSvsAU}b  for a population of PS, AS and AU corroborate this analytical condition. Equation \ref{eq:risk_dom_social_pun}  splits the region (\textbf{II})  into two parts, (\textbf{IIa}) and (\textbf{IIb}),  where PS is now also be preferred to AU in the first one. In part (\textbf{IIa}),  the transition is stronger from AU to PS than vice versa (see Figure \ref{fig:panel_markov}b).  Recall that in the whole region (\textbf{II}) the transition is stronger from AS to AU, thus leading to a cyclic pattern between these three strategies. 

When relaxing the assumption that $s_\beta = s_\alpha$ {(see SI for the detailed calculation of payoffs)}, the effect of punishment for all variations of the parameters can be studied. The results are shown in  Figure \ref{fig:reward_vs_punishment_countour_vary_sAlpha_sBeta} (bottom row), for all the three regions shown in Figure \ref{fig:reward_vs_punishment_countour_vary_sAlpha_sBeta}  in inverse order. First, when looking at the right panel (bottom row) of Figure \ref{fig:reward_vs_punishment_countour_vary_sAlpha_sBeta}, one can observe that punishment does not alter the desired outcome (safety behaviour is the preferred outcome) in region (\textbf{I}), i.e. safe behaviour remains dominant. Significant less unsafe behaviour is observed in region (\textbf{II}) , i.e. the middle panel (bottom row) of Figure \ref{fig:reward_vs_punishment_countour_vary_sAlpha_sBeta}, where it is not desirable, especially when $s_\alpha$ is small and $s_\beta$ is sufficiently large (purple area). However, punishment has an undesirable effect in region (\textbf{III}), i.e. the left panel (bottom row) of Figure \ref{fig:reward_vs_punishment_countour_vary_sAlpha_sBeta},  as it leads to reduction of AU when punishment is highly efficient (see the non-red area) while AU remains the preferred collective outcome in that region. The reason is that, for sufficiently small $s_\alpha$ and large $s_\beta$ (such that  $s^\prime > 0$ and $s^\prime > s^{\prime\prime}$), PS  gains significant advantage against AU, thereby dominating it even for low $p_r$. 


In summary, reducing the development speed of unsafe players leads to a positive effect, especially when the personal cost is much less than the effect it induces on the unsafe player.  Yet at the same time, it may lead to unwanted sanctioning effects in the region where risk-taking should be promoted.

\subsection*{Reward vs punishment for promoting safety compliance}
Here we investigate how positive incentives, as explained in Methods, influence the outcome in all three regions.  The payoff matrix showing   average payoffs among three strategies AS, AU and RS reads 
{\small
\begin{equation}
 \bordermatrix{~ & \textit{AS} & \textit{AU} & \textit{RS} \cr
                  \textit{AS} & \frac{B}{2W} +\Pi_{11} & \Pi_{12}   &  \frac{B (1 + s_\beta)}{W}+\Pi_{11} \cr
                  \textit{AU} &  p \left(\frac{sB}{W} + \Pi_{21}\right)   &  p \left(\frac{sB}{2W} +\Pi_{22}\right)  &  p \left(\frac{sB}{W} + \Pi_{21}\right)  \cr 
                  \textit{RS} & \Pi_{11}   & \Pi_{12} &  \frac{B (1 + s_\beta - s_\alpha)}{2W} +\Pi_{11}    \cr
                 }.
\end{equation}}
The payoff of RS against another RS is given under the assumption that reward is sufficiently cost-efficient, such that $1 + s_\beta >  s_\alpha$; otherwise, this payoff would be  $\Pi_{11}$. On the one hand, one can observe that RS is always dominated by AS.  On the other hand, the condition for RS to be risk-dominant against AU is  given by: 
$$ p \left(\frac{sB}{2W} +\Pi_{22} + \frac{sB}{W} + \Pi_{21}\right) <  \Pi_{12} +  \frac{B (1 + s_\beta - s_\alpha)}{2W} +\Pi_{11},$$
which, for sufficiently large $B$ (fixing $W$),  is equivalent to 
\begin{equation} 
\label{eq:Reward_riskdom_AU}
p_r > 1 - \frac{1+s_\beta - s_\alpha}{3s}.
\end{equation}
Hence, RS can improve upon AS when playing against AU whenever $s_\beta > s_\alpha$ (recall that the condition for AS to be risk-dominant against AU is $p_r > 1 - 1/(3s)$). It is different from the peer punishment strategy PS that can lead to improvement even when $s_\beta \leq s_\alpha$. 

{Thus, under the above condition, a cyclic pattern emerges (see Figure \ref{fig:panel_markov}b): from AS to AU, to RS, then back to AS. In contrast to punishment, the rewarding strategy RS has a very low frequency in general (as it is always dominated by the non-rewarding safe player AS). Nonetheless, RS catalyses the emergence of safe behaviour.  }

Figure \ref{fig:reward_vs_punishment_countour_vary_sAlpha_sBeta} (top row) shows the frequencies of AU in a population with AS and RS, for varying $s_\alpha$ and $s_\beta$, in comparison with those from the punishment model, for the three regions. One can observe that, in region (\textbf{II}), i.e. the middle panel (top row) of Figure  \ref{fig:reward_vs_punishment_countour_vary_sAlpha_sBeta}, punishment is more (or at least as) efficient than reward in suppressing AU except  
for when incentivising is rather costly (i.e. sufficiently large $s_\alpha$) but highly cost-efficient  ($s_\beta > s_\alpha$) (the areas inside the white triangles;  see also  Figure \ref{fig:reward_vs_punishment_countour_vary_sAlpha_sBeta_LargeBeta} in SI for clearer difference with larger $\beta$). It is because only when incentive is highly cost-efficient, RS can take over AU effectively (see again Equation \ref{eq:Reward_riskdom_AU}); and furthermore, the larger both $s_\alpha$ and $s_\beta$ are, the stronger the transition from RS to AS, to a degree that can overcome the transition from AS to AU. For an example satisfying these conditions, where $s_\alpha = 1.5$ and $s_\beta = 3.0$, see Figure \ref{fig:panel_markov_reward_better_punishment} in SI. 

In regions (\textbf{I}) and (\textbf{III}),  i.e. the right and left panels (top row) of Figure \ref{fig:reward_vs_punishment_countour_vary_sAlpha_sBeta},  {similarly to punishment,} the rewarding strategy does not change the  outcomes, as is desired. Note however that differently from punishment, in region (\textbf{I}), i.e. the right panel (top row) of Figure \ref{fig:reward_vs_punishment_countour_vary_sAlpha_sBeta},  only AS  dominates the  population, while in the case of punishment,  AS and PS are neutral and together dominate the population (see Figure \ref{fig:reward_vs_punishment_countour_vary_sAlpha_sBeta}, comparing panels c and f).  Most interestingly, rewards do not harm region  (\textbf{III}), i.e. the left panel (top row) of Figure \ref{fig:reward_vs_punishment_countour_vary_sAlpha_sBeta}, which suffers from over-regulation in the case of punishment because of the stronger transitions from RS to AS and AS to AU. Additional numerical  analysis shows that all these observations  are robust for larger $\beta$ (see SI, Figure \ref{fig:reward_vs_punishment_countour_vary_sAlpha_sBeta_LargeBeta}).

In SI, we also consider the scenario where both peer reward and punishment are present, in a population of four strategies, AS, AU PS and RS (see Figures \ref{fig:panel_markov_four_strategies} and \ref{fig:reward_and_punishment_countour_vary_sAlpha_sBeta}). Since PS behaves in the same way as AS when interacting with RS, there is always a stronger transition from RS to PS. It results in an outcome in terms of AU frequency similar to the case when only PS is present, suggesting that, in a self-organized scenario, peer-punishment is more likely to prevail than peer-rewarding when individuals face a technological race.

 {Finally, it is noteworthy that all results obtained in this paper are robust if one considers that with some probability in each round UNSAFE players can be detected resulting in those UNSAFE players losing all payoff in that round   \cite{han2019modelling}.  This observation confirms  the observation in that in a short-term AI regime only participants' speeds matter (in relation to the disaster risk, $p_r$), and controlling the speeds is important to ensure a beneficial outcome (see also \cite{han2019modelling}).   }

\section*{Conclusion}

In this paper we study the dynamics associated with technological races, those having the objective of being the first to bring some AI technology to market as a case study. The model proposed, however, is general enough for applicability to other innovation dynamics which face the conflict between safety and rapid development \cite{denicolo2010winner,abbott2009global}.   We address this problem resorting to a multiagent and complex systems approach, while adopting well established methods from evolutionary game theory and populations dynamics . 


We propose a plausible adaptation of a baseline model \cite{han2019modelling} which can be useful when thinking about policies and regulations, namely incipient forms of community enforcing mechanisms, such as peer rewards and sanctions. We identify the conditions under which these  incentives provide the desired effects while highlighting the importance of {clarifying the risk disaster regimes and} the time-scales  associated with the problem. In particular, our results suggest that punishment --- by forcibly reducing the development speed of unsafe participants --- can generally reduce unsafe behaviour even when sanctions are not particularly efficient. In contrast, when punishment is highly efficient, it can lead to over-regulation and an undesired reduction of innovation, noting that  a speedy and unsafe development is acceptable and more beneficial for the whole population whenever the risk for setbacks or disaster is low compared to the extra speed gained by ignoring safety precautions.  Similarly, rewarding a safe co-player to speed up its development may, in some regimes, stimulate safe behaviours, whilst avoiding the detrimental impact of over-regulation. 

These results show that, similarly to peer incentives in the context of one-shot social dilemmas  (such as the Prisoner's Dilemma and the Public Goods Game) \cite{key:fehr2002,Sigmund2001PNAS,boyd2010coordinated,sigmundinstitutions,hilbe2012emergence,szolnoki2013correlation,gois2019reward,Hanijcai2018,chen2015first,garcia2019evolution}, strategies that target development speed in DSAIR can influence the evolutionary dynamics, but interestingly, they produce some very different effects from those of incentives in social dilemmas \cite{perc2017statistical}. For example, we have shown  that strong punishment, even when highly inefficient, can lead to  improvement of safety outcome; while punishment in social dilemmas can promote cooperation only when highly cost-efficient.  On the other hand, when punishment is too strong, it might lead to an undesired effect of over-regulation (reducing innovation where desirable), which is not generally the case in social dilemmas.

Our model and analysis of elementary forms of incentives thus  provides  an instrument for policy makers to
 ponder on the supporting mechanisms (e.g. positive and  negative incentives), in the context of technological races \cite{sotala2014responses,burrell2014public,brundage2020toward}. 
Concretely, both sanctioning of wrong-doers (e.g. rogue or unsafe developers/teams) and rewarding of right-doers (e.g. safe-compliant developers/teams) can lead to enhancement of the desirable outcome (it being that of innovation or risk-taking in  low risk cases, and safety-compliance in higher risk cases). Notably, while the former can  be detrimental for innovation in low risk cases, it leads to a stronger enhancement for a wider range of  effect-to-cost ratio of incentives.  
Thus, when it is not clear from the beginning what is the risk level associated (with the technology to be developed),  then positive incentives  appear to  be the safer choice than negative ones (in line with historical data on rewards usage in  innovation policy in the UK \cite{burrell2014public} as well as suggestions  for Covid-19 vaccine innovation policy  \cite{burrell2020covid}). This is the case for many kinds of technological races especially when data about the  effect of a new technology is usually  lacking and  only becomes available when  it has been created and used enough (see the Collingridge Dilemma \cite{Collingridge1980}), as are the cases of the domain supremacy race through AI \cite{armstrong2014errors,grace2018will} and the race for creating the first Covid-19 vaccines \cite{callaway2020race,burrell2020covid}.   On the other hand, when one can determine early on that the associated level of risk is sufficiently high (i.e. above a certain threshold as determined in our analysis), negative incentives might provide a stronger mechanism. For instance, high risk technologies such as new airplane models  and medical products \cite{world2003medical} might benefit from  putting strong sanctioning mechanisms in place.

In short, our analysis has shown, within an idealised model of an AI race and using a game theoretical framework, that some simple forms of peer incentives, if used suitably (to avoid over-regulation, for example) can provide a way to escape the dilemma of acting safely even when speedy unsafe development is preferred. Future studies may look at more complex incentivising mechanisms \cite{brundage2020toward} such as reputation and public image manipulation \cite{santos2018social,santos2018indirect}, emotional motives of  guilt and apology-forgiveness  \cite{Luis2017AAMAS,martinez2015apology}, institutional and coordinated incentives \cite{sigmundinstitutions,vasconcelos2013bottom}, and the subtle combination of different forms of incentive  (e.g., stick-and-carrot approach and incentives for  agreement compliance) \cite{han2018cost,hanTom2016synergy,chen2015first,gois2019reward,wang2019exploring}.   

\section*{Acknowledgements}

T.A.H., L.M.P. and T.L. have been supported by Future of Life Institute grant RFP2-154. L.M.P. also acknowledges support from FCT/MEC NOVA LINCS PEst UID/CEC/04516/2019. F.C.S. acknowledges support from FCT Portugal (grants PTDC/EEI-SII/5081/2014, PTDC/MAT/STA/3358/2014, and UID/CEC/50021/2020). T.L. acknowledges support by the FuturICT2.0 (www.futurict2.eu) project funded by the FLAG-ERA JCT 2016.

\section*{Supporting information}

\subsection*{Details of analysis for three strategies AS, AU, CS}
Let CS be a conditionally safe strategy, playing SAFE in the first round and choosing the same move as the co-player's choice in the previous round. {We recall below the detailed calculations for this case, as described in \cite{han2019modelling}, just  for completeness.}  
The average payoff matrix for the three strategies AS, AU, CS reads (for row player) 
{\small
\begin{equation}
\centering
\Pi = \bordermatrix{~ & \textit{AS} & \textit{AU} & \textit{CS} \cr
                  \textit{AS} & \frac{B}{2W} +\pi_{11} & \pi_{12}   &  \frac{B}{2W}+\pi_{11} \cr
                  \textit{AU} &  (1 - p_r) \left(\frac{sB}{W} + \pi_{21}\right)   &  (1 - p_r) \left(\frac{sB}{2W} +\pi_{22}\right)  &  (1 - p_r)\left[\frac{sB}{W} +  \frac{s }{W}  \left(\pi_{21} + (\frac{W}{s} - 1) \pi_{22}  \right)\right] \cr 
                  \textit{CS} &  \frac{B}{2W} +\pi_{11}    & \frac{s}{W}  \left( \pi_{12} + (\frac{W}{s} - 1) \pi_{22}  \right) &  \frac{B}{2W} +\pi_{11}    \cr
                 }.
                \end{equation}
}
The conditions (i) SAFE population has a larger average payoff than that of UNSAFE one, i.e.  $\Pi_{AS,AS} > \Pi_{AU, AU}${, meaning by definition that a collective outcome is preferred} and (ii) when is it the case that AS and CS are more likely to be imitated against AU   (i.e., risk-dominant) will be derived below. First, for condition  (i), it must hold that 

\begin{equation}
\label{eq:population_social_welfare1}
\frac{B}{2W} +\pi_{11} > (1-p_r) \left(\frac{sB}{2W} + \pi_{22}\right).
\end{equation}
Thus, 
\begin{equation}
\label{eq:population_social_welfare}
 p_r >1 - \frac{B  + 2 W \pi_{11} }{ s B + 2 W \pi_{22} },
\end{equation}
which is equivalent to (since $B/W \gg b$)
\begin{equation} 
\label{eq:safety_prefer}
 p_r > 1 - \frac{1}{s}. \ \quad 
\end{equation}
   {This inequality means that,} whenever the risk  of a    {disaster or personal setback, $p_r$,}  is larger than  the gain that can be gotten from a greater development speed, then the preferred collective action in the population is safe{ty compliance}.  

   {Now, for  condition (ii),} 
\begin{equation}
\label{eq:risk_dom_AUvsCSAS_SI}
\frac{B}{2W}  + \pi_{11} + \pi_{12} > (1-p_r) \left(\frac{3sB}{2W} + \pi_{21} + \pi_{22}\right).
\end{equation}
\begin{equation}
\label{eq:risk_dom_AUvsCSAS_SI}
\begin{split}
 &\frac{s}{W}  \left( \pi_{12} + (\frac{W}{s} - 1) \pi_{22}  \right) + \frac{B}{2W} +\pi_{11}  \\  &>  (1 - p_r) \left[\frac{sB}{2W} + \frac{sB}{W} +  \frac{s }{W}  \left(\pi_{21} + (\frac{W}{s} - 1) \pi_{22}  \right) + \pi_{22} \right],
 \end{split}
\end{equation}
which are both equivalent to (since $B/W \gg b$)

\begin{equation}
\label{eq:risk_dom_AUvsCSAS}
p_r > 1-\frac{1}{3s}.
\end{equation}
The two boundary conditions    {for (i) and (ii), as given in Equations  \ref{eq:safety_prefer} and \ref{eq:risk_dom_AUvsCSAS},     {splits} $s - p_r$ parameter space} into  three regions, as exhibited in Figure \ref{fig:panel_no_punishmentSI}a: 
\begin{itemize}
\item[(\textbf{I})]  when $ p_r > 1-\frac{1}{3s}$:  This corresponds to the \emph{AIS compliance zone}, in which safe AI    {compliance is both  preferred collectively and that  unconditionally (AS) and conditionally (CS) safe development}  is the social norm ({an example for $s = 1.5$ is given in   Figure \ref{fig:panel_no_punishmentSI}b:} $p_r > 0.78$); 
\item[(\textbf{II})] when $1-\frac{1}{3s} > p_r > 1-\frac{1}{s}$: This intermediate zone is the one that captures a dilemma because, collectively, safe AI developments are preferred, though the social dynamics pushes the whole population to the state where all develop AI in an unsafe manner. We shall refer to this zone as the \emph{AIS dilemma zone} ({for $s = 1.5$, $0.78 > p_r > 0.33$, see Figure \ref{fig:panel_no_punishmentSI}c});  
\item[(\textbf{III})] when $p_r <  1-\frac{1}{s}$: This defines the \emph{AIS innovation zone}, in which  unsafe    {development is not only the preferred collective outcome but also} the one the social dynamics selects.  
\end{itemize}

\subsection*{{Calculation for $\pi_{PS, AU}$  and  $\pi_{AU, PS}$ in general case}}
Below $R$ denotes the average number of rounds; $B_1$ and $B_2$ the benefits PS and AU might obtain from the winning benefit $B$ when either of them wins the race by being the first to have made $W$ development steps;  $b_1$ and $b_2$  the intermediate benefits  PS and AU might obtain in each round of the game; $p_{loss}$ is the probability that all the benefit is not lost  when AU wins  and draws the race;  Clearly, all these values depend on the development speeds ($s^\prime$ for PS and $s^{\prime\prime}$ for AU). 

 $$\pi_{PS \ vs \ AU} = \frac{1}{R(s^\prime, s^{\prime\prime})}  \left[ \pi_{12} + B_1(s^\prime, s^{\prime\prime}) + (R(s^\prime, s^{\prime\prime})-1) (-c + b_1(s^\prime, s^{\prime\prime}))  \right] $$
 
 $$\pi_{PS \ vs \ AU} = p_{loss}(s^\prime, s^{\prime\prime})\times\frac{1}{R(s^\prime, s^{\prime\prime})}  \left[ \pi_{21} + B_2(s^\prime, s^{\prime\prime}) + (R(s^\prime, s^{\prime\prime})-1)  b_2(s^\prime, s^{\prime\prime})  \right] $$
 
 where $ B_1(s^\prime, s^{\prime\prime})  = \begin{cases} B &\mbox{if } s^\prime > 0 \ \& \ s^{\prime\prime} \leq 0\\
 B &\mbox{if } s^\prime > 0 \ \& \ \frac{W-s}{s^{\prime\prime}} > \frac{W-1}{s^{\prime}}  \\
 B/2 &\mbox{if } s^\prime > 0 \ \& \  \frac{W-s}{s^{\prime\prime}} = \frac{W-1}{s^{\prime}}  \\
0 & \mbox{otherwise } \end{cases}  $
 
$ B_2(s^\prime, s^{\prime\prime})  = \begin{cases} B &\mbox{if } s^\prime \leq 0 \ \& \ s^{\prime\prime} > 0\\
B &\mbox{if } s^{\prime\prime} > 0 \ \& \ \frac{W-s}{s^{\prime\prime}} < \frac{W-1}{s^{\prime}}  \\
 B/2 &\mbox{if } s^{\prime\prime} > 0 \ \& \  \frac{W-s}{s^{\prime\prime}} = \frac{W-1}{s^{\prime}}  \\
0 & \mbox{otherwise } \end{cases}  $
 
   $ b_1(s^\prime, s^{\prime\prime})  = \begin{cases}  (1-p_{fo}) \frac{s^\prime b}{s^\prime+s^{\prime\prime}}    + p_{\mathit{fo}} b   &\mbox{if } s^\prime > 0 \ \& \ s^{\prime\prime} > 0  \\
  b   &\mbox{if } s^\prime > 0 \ \& \ s^{\prime\prime} \leq 0  \\
0 & \mbox{otherwise } \end{cases}  $

   $ b_2(s^\prime, s^{\prime\prime})  = \begin{cases}  (1-p_{fo}) \frac{s^{\prime\prime} b}{s^\prime+s^{\prime\prime}}      &\mbox{if } s^\prime > 0 \ \& \ s^{\prime\prime} > 0  \\
  (1-p_{fo}) b       &\mbox{if } s^\prime \leq 0 \ \& \ s^{\prime\prime} > 0  \\
0 & \mbox{otherwise } \end{cases}  $

 $ R(s^\prime, s^{\prime\prime})  = \begin{cases} +\infty     &\mbox{if } s^\prime \leq 0 \ \& \ s^{\prime\prime} \leq 0  \\
 \frac{W-1}{s^{\prime}}  +1  &\mbox{if } s^\prime > 0 \ \& \ s^{\prime\prime} \leq 0  \\
 \frac{W-s}{s^{\prime\prime}} +1      &\mbox{if } s^\prime \leq 0 \ \& \ s^{\prime\prime} > 0  \\
1+ \min\{\frac{W-s}{s^{\prime\prime}}, \frac{W-1}{s^{\prime}}  \} & \mbox{otherwise } \end{cases}  $

  $ p_{loss}(s^\prime, s^{\prime\prime})  = \begin{cases}  p (= 1 - p_{r})     & \mbox{if } s^{\prime\prime} > 0 \ \& \ \frac{W-s}{s^{\prime\prime}} \leq \frac{W-1}{s^{\prime}}   \\
1 & \mbox{otherwise } \end{cases}  $

\begin{figure*}
\centering
\includegraphics[width=\linewidth]{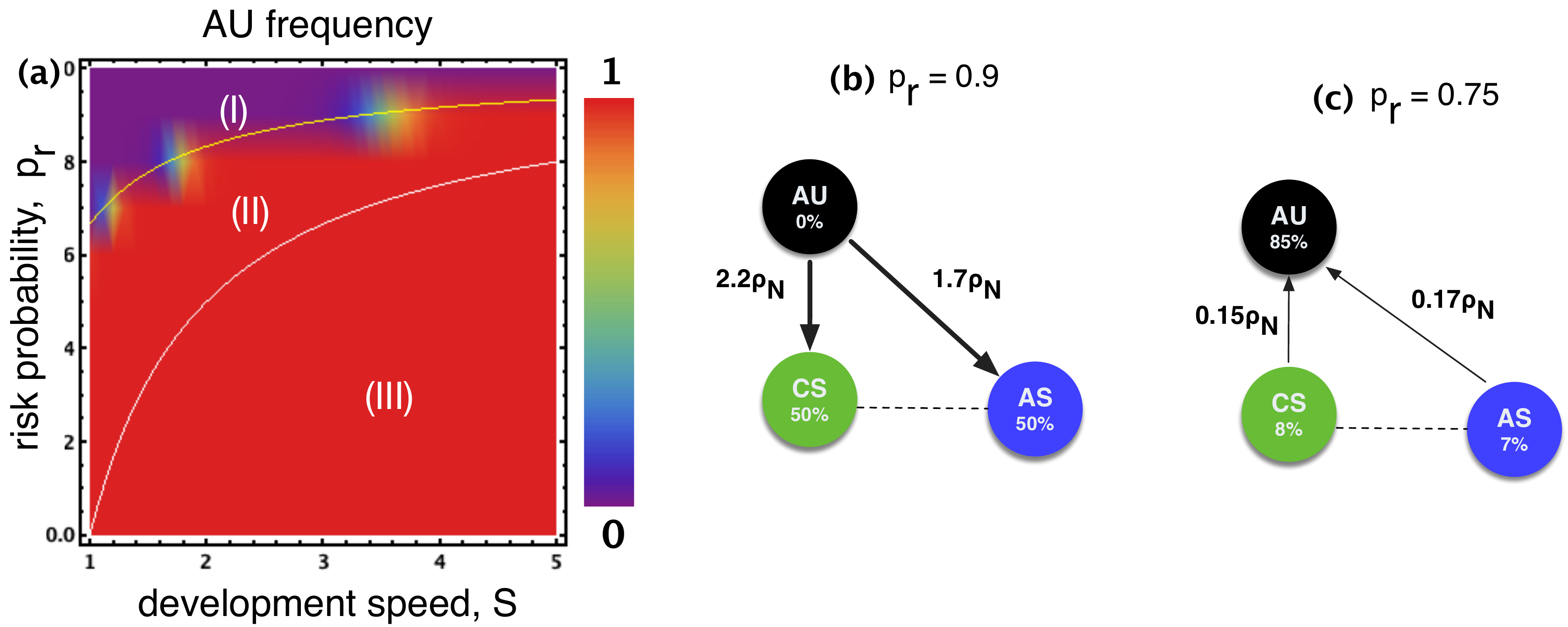}
\caption{     {Panel \textbf{(a)} as in Figure 1 in the main text, added here for ease of following.} Panels \textbf{(b)} and \textbf{(c)}    {show the transition probabilities and stationary distribution (see Methods)}.  In panel (c) AU dominates, corresponding to region (\textbf{II}), whilst in panel (b) AS and CS dominate, corresponding to region (\textbf{I}).    {For a clear presentation, we  indicate just} the stronger directions.  Parameters: $b = 4$, $c = 1$, $W = 100$,  $B = 10^4$,  $Z = 100$, $\beta = 0.1$    {; In panel \textbf{(b)} $p_r = 0.9$; in  panel (c) $p_r = 0.6$; in both (b) and (c) $s = 1.5$}. 
}
\label{fig:panel_no_punishmentSI}
\end{figure*}

\begin{figure*}
\centering
\includegraphics[width=0.8\linewidth]{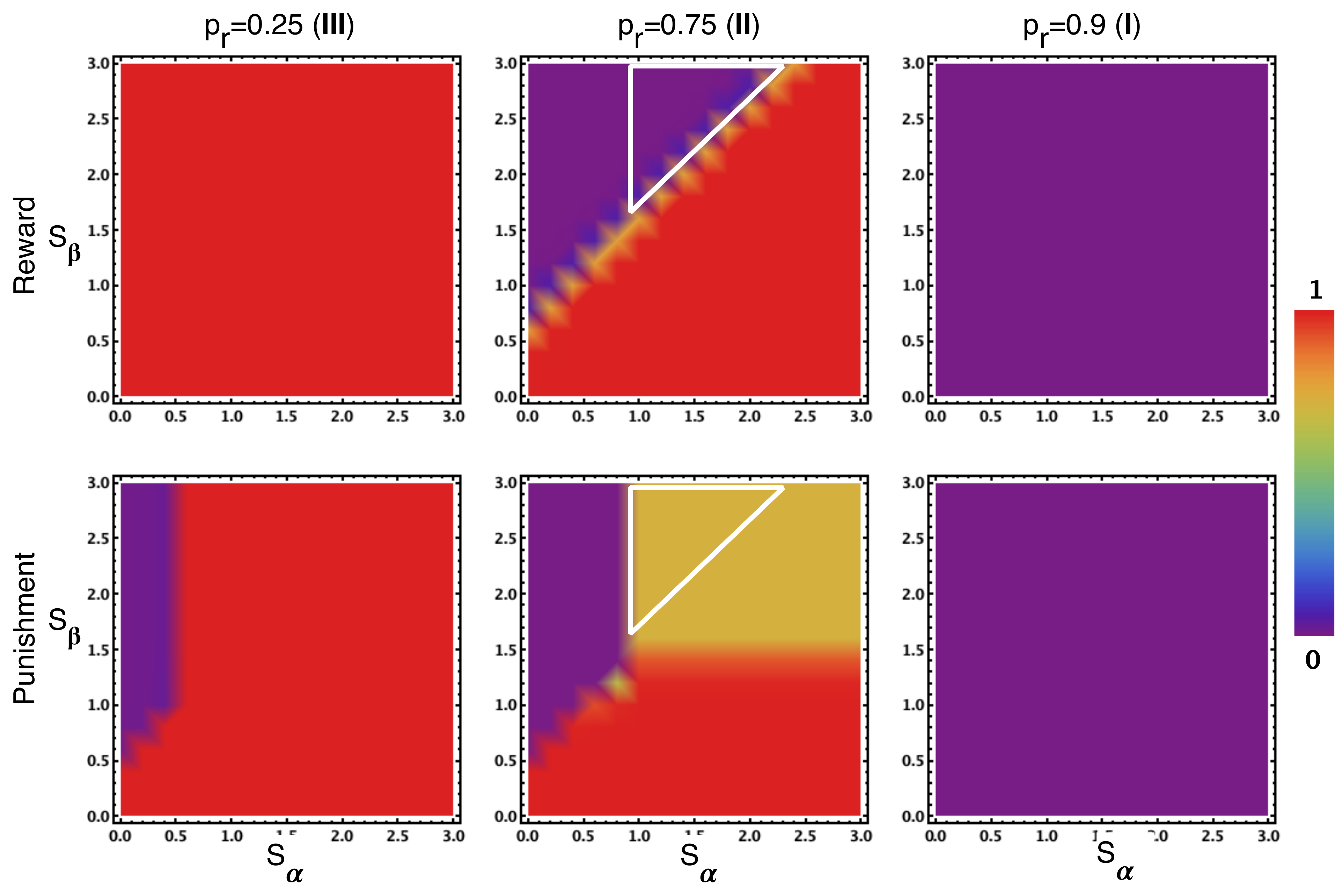}
\caption{\textbf{AU Frequency: Reward (top row) vs punishment (bottom row) for varying $s_\alpha$ and $s_\beta$}, for three regions, for stronger intensity of selection ($\beta = 0.1$).  Other parameters are the same as in Figure \ref{fig:reward_vs_punishment_countour_vary_sAlpha_sBeta} in the main text. The observations in that figure is also robust for larger intensities  of selection.   }
\label{fig:reward_vs_punishment_countour_vary_sAlpha_sBeta_LargeBeta}
\end{figure*}

\begin{figure*}
\centering
\includegraphics[width=\linewidth]{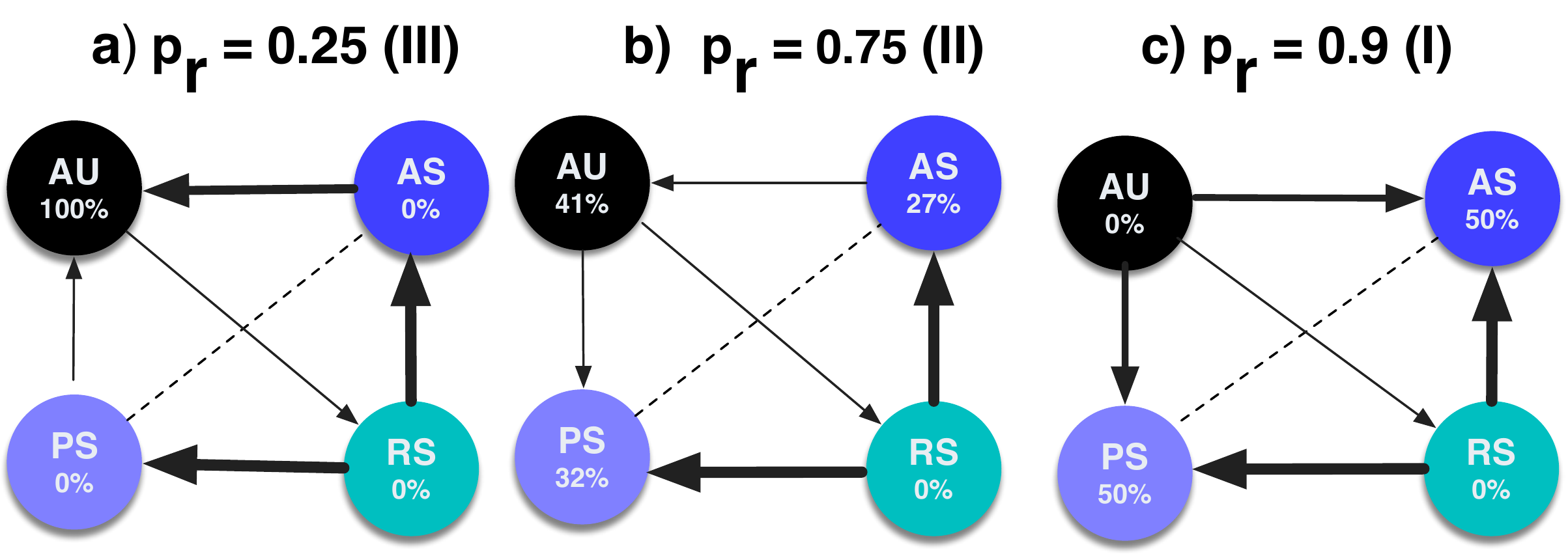}
\caption{\textbf{Transitions and stationary distributions  in a population of four strategies AU, AS, PS and RS}, for three regions. Only stronger transitions are shown for clarity. Dashed lines denote  neutral transitions. In addition, note that PS is equivalent to AS when interacting with PS, i.e. there is always a stronger transition from RS to PS than vice versa.   Parameters as in Figure 2. 
}
\label{fig:panel_markov_four_strategies}
\end{figure*}

\begin{figure*}
\centering
\includegraphics[width=\linewidth]{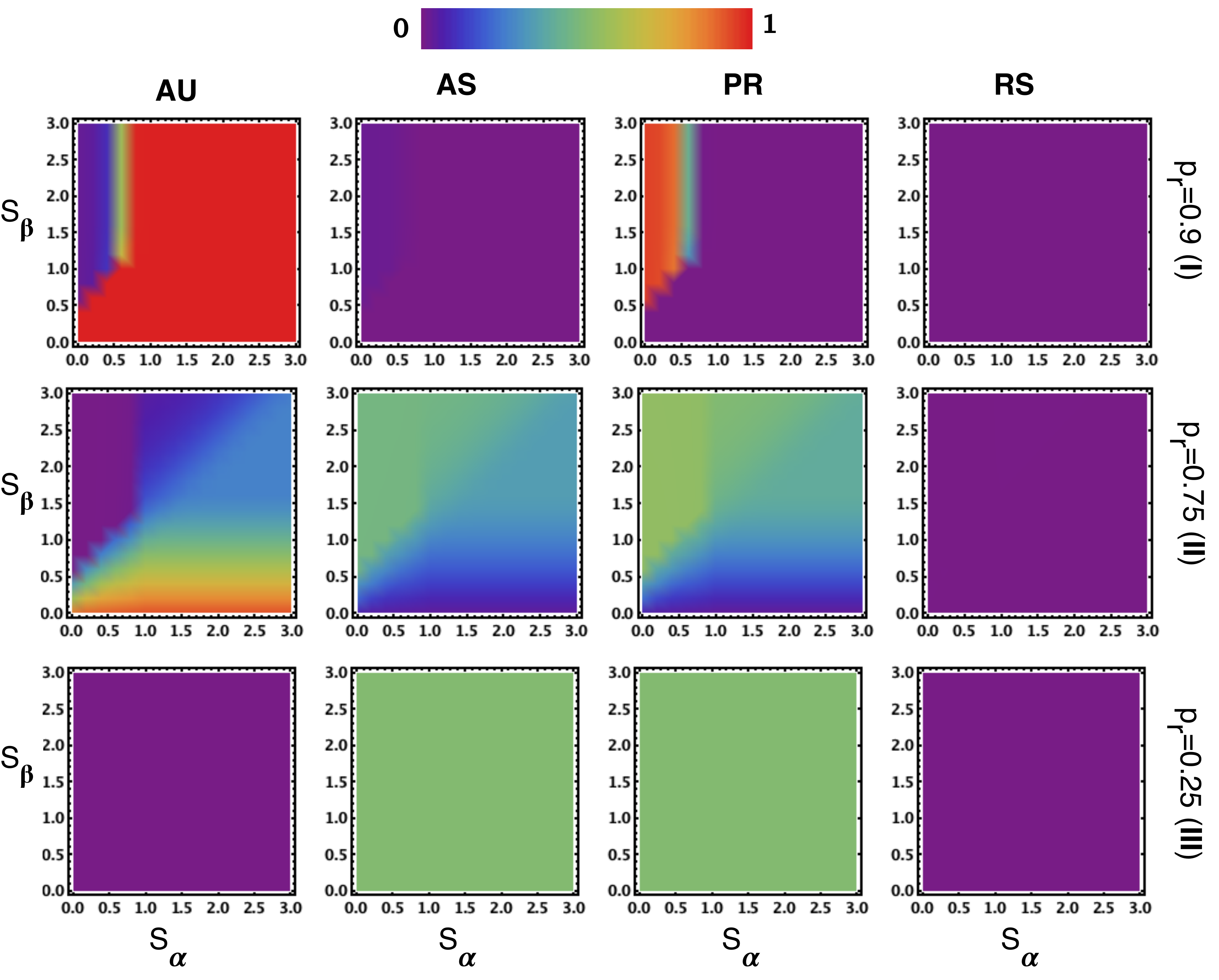}
\caption{\textbf{AU Frequency for varying $s_\alpha$ and $s_\beta$, in a population of four strategies AS, AU, PS and RS}, for three regions. The outcomes in all regions are similar to the case of punishment (without reward) in Figure  5. The reason is that there is always a stronger transition from RS to PS than vice versa.    Parameters as in Figure 5. }
\label{fig:reward_and_punishment_countour_vary_sAlpha_sBeta}
\end{figure*}

\begin{figure*}
\centering
\includegraphics[width=0.8\linewidth]{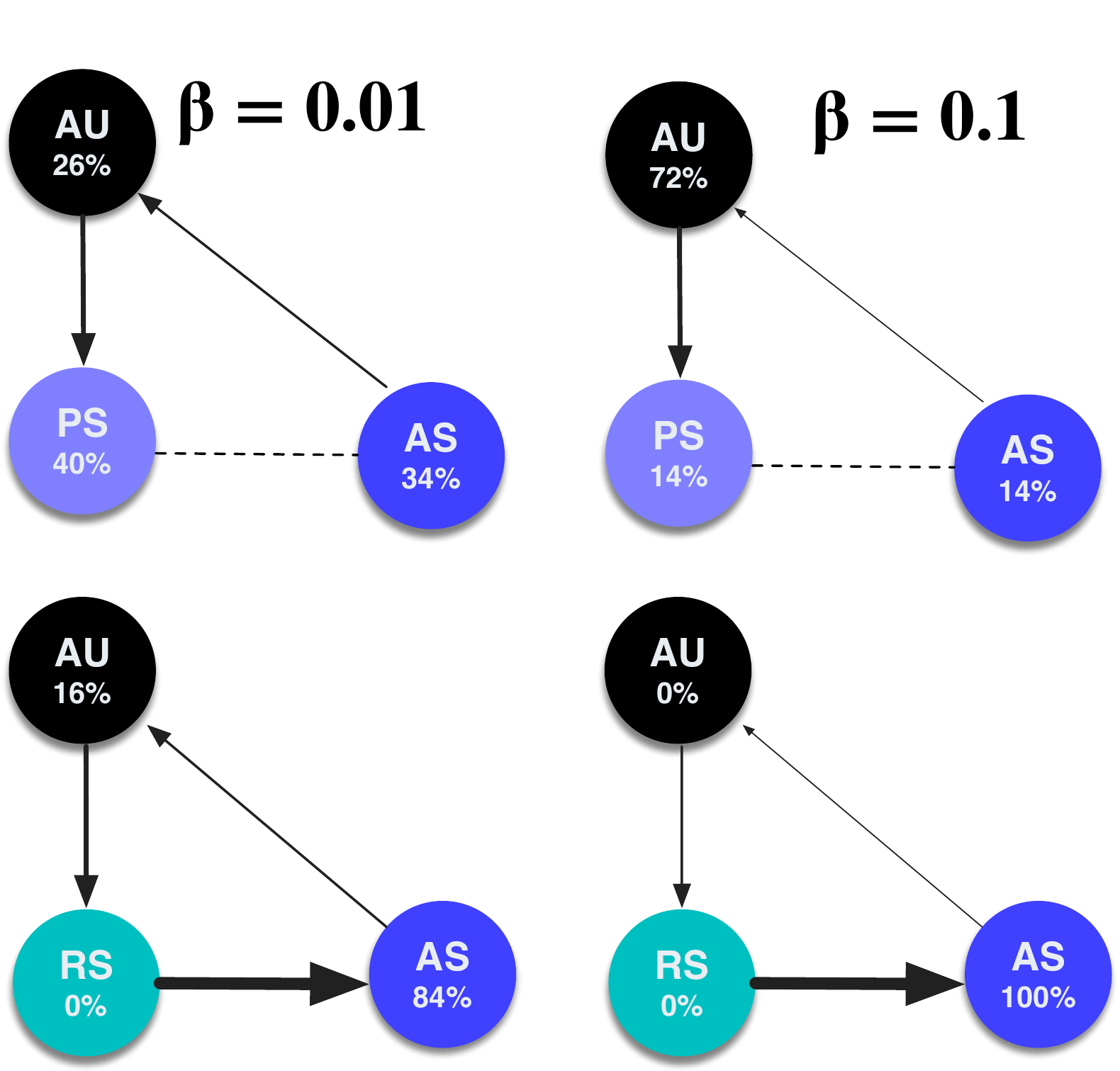}
\caption{Transitions and stationary distributions  in a population of three strategies AU, AS, with either  PS (top row) or RS (bottom row),  in region (II) ($pr = 0.75$): left column ($\beta = 0.01$), right column ($\beta = 0.1$).  The parameters of incentives fall in the white triangles in Figures 5 and 7: $s_\alpha = 1.5, \ s_\beta = 3$. We observe that the frequency of AU is lower in case of reward than that of punishment.  Other parameters as in Figure 2. 
}
\label{fig:panel_markov_reward_better_punishment}
\end{figure*}

\newpage




\end{document}